\documentclass[a4paper,fleqn]{cas-dc}

\usepackage[numbers]{natbib}
\usepackage{multirow}
\usepackage{algorithm}
\usepackage{algpseudocode}

\def\tsc#1{\csdef{#1}{\textsc{\lowercase{#1}}\xspace}}
\tsc{WGM}
\tsc{QE}
\tsc{EP}
\tsc{PMS}
\tsc{BEC}
\tsc{DE}

\begin{document}
\let\WriteBookmarks\relax
\def\floatpagepagefraction{1}
\def\textpagefraction{.001}
\shorttitle{MAPR for 3D Point Cloud Robustness}
\shortauthors{Alonso et~al.}

\title [mode = title]{Beyond Defenses: Manifold-Aligned Regularization for Intrinsic 3D Point Cloud Robustness}                      



\author[1,2]{Pedro Alonso}[type=editor,
                        auid=000,bioid=1,
                        orcid=0009-0005-0857-8821]
\ead{palonso@swjtu.edu.cn}

\credit{Conceptualization, Methodology, Software, Validation, Formal analysis, Writing– original draft, Visualization, Funding acquisition}

\affiliation[1]{organization={School of Computing and Artificial Intelligence, Southwest Jiaotong University},
                city={Chengdu},
                postcode={611756}, 
                country={China}}

\affiliation[2]{organization={Engineering Research Center of Sustainable Urban Intelligent Transportation, Ministry of Education, Southwest Jiaotong University},
                city={Chengdu},
                postcode={611756}, 
                country={China}}

\author[1,2]{Chongshou Li}[type=editor,
                        auid=000,bioid=2,
                        orcid=0000-0002-7595-0997]
\cormark[1]
\ead{lics@swjtu.edu.cn}

\credit{Supervision, Funding acquisition}

\author[1,2]{Tianrui Li}[type=editor,
                        auid=000,bioid=3,
                        orcid=0000-0001-7780-104X]
\ead{trli@swjtu.edu.cn}

\credit{Supervision}

\cortext[cor1]{Corresponding author}


\begin{abstract}
Despite extensive progress in point cloud robustness, existing methods primarily rely on augmentation strategies or defense mechanisms while overlooking the geometric nature of adversarial fragility. We hypothesize that adversarial vulnerability in 3D networks arises from a manifold misalignment between the latent geometry learned by the model and the intrinsic geometry of the underlying surface. Small, geometry-preserving perturbations along the input manifold often induce disproportionate distortions in feature space, potentially leading to misclassifications. We formalize this phenomenon by developing a geometric interpretation of 3D robustness that links classical adversarial theory to the intrinsic structure of point clouds. Motivated by this analysis, we introduce Manifold-Aligned Point Recognition (MAPR), a framework that regularizes the latent geometry by aligning predictions across intrinsic perturbations. MAPR augments each point cloud with intrinsic features capturing local curvature and diffusion structure, and applies a consistency loss that preserves invariance to intrinsic, geometry-preserving perturbations. Without relying on adversarial training or additional data, MAPR consistently improves robustness under multiple adversarial attacks across several datasets, achieving average robustness gains of +20.02 and +8.83 percentage points over vanilla models on ModelNet40 and ScanObjectNN, respectively.
\end{abstract}




\maketitle

\section{Introduction}
\label{sec:intro}

Deep neural networks (DNNs) are well known to be vulnerable to adversarial attacks \cite{szegedy2014intriguing, Carlini2016TowardsET, 8803770, 8953645, Zhang_Gu_Huang_Jiang_Wu_Lyu_2024, CHEN2024103539}, which introduce small, often imperceptible perturbations capable of misleading models into incorrect predictions. Although robustness against adversarial attacks has been extensively studied in 2D images, it remains particularly challenging in 3D point clouds due to their unordered structure, local sparsity, and geometric variability. Despite steady progress in defense techniques, point cloud classifiers remain highly fragile: small changes in sampling density, local neighborhood structure, or viewpoint can drastically alter predictions. Existing approaches based on outlier removal \cite{DBLP:conf/iccv/ZhouCZFZY19, Li_Lu_Ding_Sun_Zhou_Chee_2024}, adversarial training \cite{8803770, 2019arXiv190710764Z, 10.1109/TIP.2024.3372456, 10.5555/3737916.3740093}, data augmentation \cite{10.1007/978-3-030-58580-8_20, 9577799}, and input purification or restoration \cite{2020arXiv201005272W, Sun2023ACR} have improved robustness, but often require additional preprocessing, attack-specific training, or substantial computational overhead. Despite their effectiveness, most existing defenses operate primarily in the input space, rather than explicitly enforcing robustness in the geometry of the learned latent representations. This observation suggests that adversarial fragility in point cloud networks may be fundamentally related to how intrinsic geometric structure is represented in latent space.

\begin{figure}
  \centering
  \includegraphics[width=\linewidth]{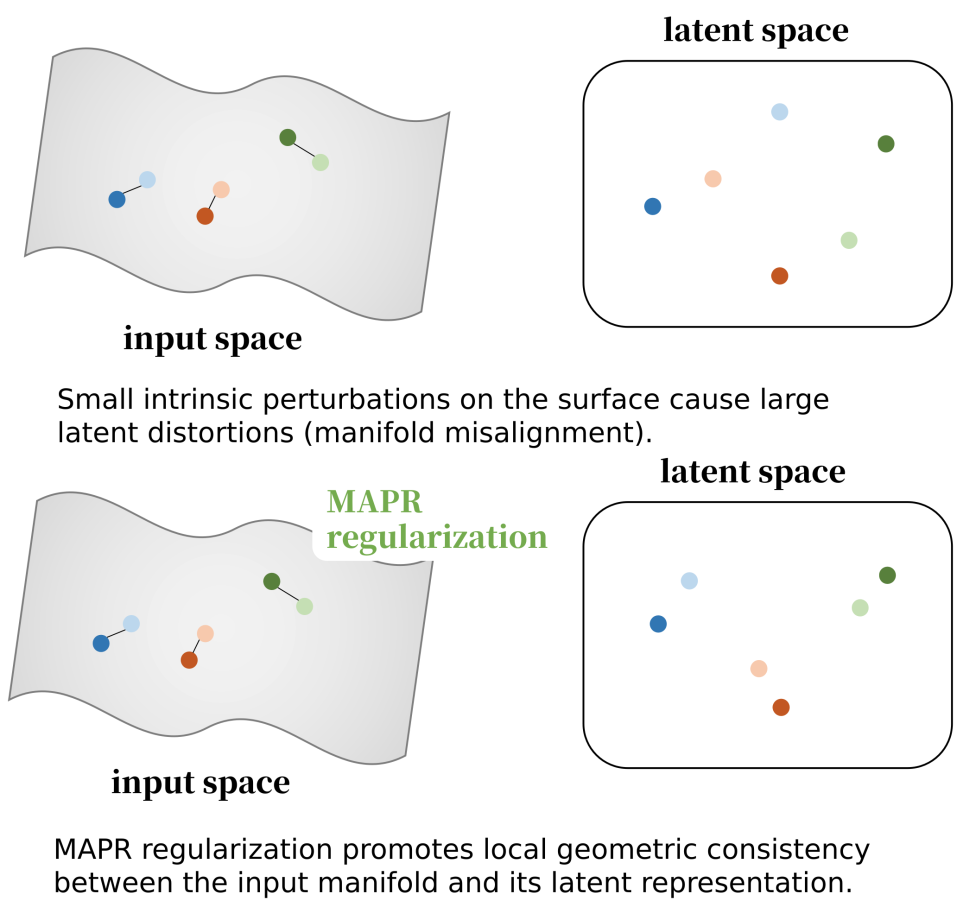}
  \caption{Illustration of manifold misalignment and the effect of MAPR. Intrinsic perturbations in the input manifold can be amplified in latent space, while MAPR regularization encourages local consistency between input neighborhoods and their latent representations.}
  \label{fig:plot_visual}
\end{figure}

From this perspective, adversarial fragility can be interpreted geometrically. Point clouds lie on a low-dimensional manifold embedded in $\mathbb{R}^3$, where intrinsic structure, captured by geodesic distances, curvature, and surface continuity, reflects the true geometry of the underlying object. We hypothesize that adversarial vulnerability in 3D arises from a mismatch between the intrinsic geometry of the object and the latent geometry learned by the network. Such misalignment naturally emerges in existing point cloud architectures. Deep networks such as PointNet \citep{Qi_2017_CVPR}, PointNet++ \citep{NIPS2017_d8bf84be}, and DGCNN \citep{10.1145/3326362} construct latent feature spaces whose geometry emerges from learned similarity metrics, dynamic \emph{k}-NN graphs, and hierarchical pooling operations. When the geometry of this latent space becomes misaligned with the intrinsic manifold, the network begins to treat sampling noise and density fluctuations as discriminative features, even though they are irrelevant to the underlying shape, making the model more vulnerable to adversarial attacks.

To address this issue, we introduce Manifold-Aligned Point Recognition (MAPR), a framework built on two complementary components: (\textit{i}) intrinsic feature augmentation, which enriches each point with local geometric descriptors such as curvature and diffusion structure, and (\textit{ii}) intrinsic consistency regularization, which encourages consistent predictions under small geometry-preserving perturbations of the input. Together, these components promote local invariance to geometry-preserving perturbations and encourage the network to learn representations that better reflect the underlying geometry of the object. Unlike adversarial defenses that target specific attacks, MAPR acts as a general training regularizer that improves robustness while remaining compatible with standard defense mechanisms.

The main contributions of this work are as follows:
\begin{itemize}
\item We provide a unified geometric interpretation of adversarial fragility in 3D, linking classical adversarial theory to the geometric structure of point clouds.
\item We introduce Manifold-Aligned Point Recognition (MAPR), a regularization framework that augments point clouds with intrinsic geometric features and applies an intrinsic consistency loss to promote invariance to geometry-preserving perturbations, without requiring adversarial training or additional data.
\item We evaluate MAPR on both the ModelNet40 \citep{Wu_2015_CVPR} and ScanObjectNN \citep{9009007} datasets across five representative point cloud architectures under diverse adversarial attacks. MAPR consistently improves robustness over vanilla baselines, achieving average robustness gains of +20.02 and +8.83 percentage points on ModelNet40 and ScanObjectNN, respectively.
\end{itemize}

\section{Related Work}
\label{sec:related_work}
\subsection{Theoretical Perspectives on Adversarial Vulnerability}
The vulnerability of deep neural networks (DNNs) to adversarial examples was first identified by \cite{szegedy2014intriguing}, who showed that imperceptibly small perturbations can drastically change model predictions. This phenomenon was initially attributed to factors such as excessive network nonlinearity, overfitting, or insufficient regularization. Following this, \cite{goodfellow2014explaining} proposed the linear hypothesis, arguing that adversarial vulnerability is largely a consequence of the approximately linear behavior of modern networks in high-dimensional input spaces. In such spaces, even imperceptibly small perturbations distributed across many dimensions can accumulate and produce large shifts in the model’s output. Although activation functions such as ReLU introduce global nonlinearity, networks behave almost linearly in local regions of the input space, making them inherently fragile in high dimensions \cite{goodfellow2014explaining, 10.5555/2969033.2969153, Fawzi2015AnalysisOC}.

More recently, \cite{NEURIPS2019_e2c420d9} argued that adversarial fragility arises from the presence of non-robust features in the data. These features, although often imperceptible to humans, are genuine statistical signals that are highly predictive for classification yet inherently unstable under small perturbations. Because standard training encourages models to exploit any feature that improves accuracy, regardless of its robustness to perturbations, networks tend to rely heavily on such non-robust features, making them more fragile against adversarial attacks.

The same principles extend naturally to 3D point clouds. A point cloud with $N$ points lies in $\mathbb{R}^{3N}$, which represents a high-dimensional input space since point clouds often contain thousands of points. As a result, the linear hypothesis remains relevant: small perturbations applied over many points can accumulate and cause large shifts in the learned feature space. Moreover, architectures such as PointNet \cite{Qi_2017_CVPR}, PointNet++ \cite{NIPS2017_d8bf84be}, and DGCNN \cite{10.1145/3326362} rely heavily on max pooling and local neighborhood aggregation. As a result, their latent representations can become highly sensitive to a small number of critical points, making them particularly vulnerable to perturbations affecting those points. In addition, 3D point clouds inherently contain non-robust features, such as sampling density or viewpoint variations, that models often exploit for prediction despite not altering the underlying shape, further amplifying their adversarial vulnerability.

\subsection{Point-Based Deep Learning for 3D Classification}
Point-based classification methods operate directly on raw point cloud data using permutation-invariant architectures. The pioneering PointNet \cite{Qi_2017_CVPR} introduced the first permutation-invariant network that could process point clouds, based on shared MLPs followed by global max-pooling. Later, PointNet++ \cite{NIPS2017_d8bf84be} extended this idea with hierarchical local feature extraction, while DGCNN \cite{10.1145/3326362} introduced dynamic graph convolutions that capture local topology through edge features. More recent methods such as CurveNet \cite{curvenet2021} and PointMLP \cite{DBLP:conf/iclr/MaQYR022} improved representation capacity and efficiency by refining local aggregation mechanisms.

Despite their success, these models are still susceptible to adversarial attacks. For instance, the use of max-pooling makes models heavily dependent on a few critical points and, if an attack perturbs these points, the model's output can shift significantly. Graph-based architectures such as DGCNN dynamically update graph connections, meaning that even minor coordinate shifts can alter local topology. As a result, these networks can overfit to non-robust features such as sampling density or viewpoint.

\subsection{Adversarial Attacks and Defenses in 3D}
Adversarial attacks introduce small perturbations, often imperceptible to humans, into the input data with the goal of misleading classifiers into incorrect predictions. Mitigating such adversarial fragility is crucial for safety-critical applications such as autonomous driving, robotics, and 3D scene understanding, where model reliability under geometric perturbations is essential.

In recent years, numerous adversarial attacks have been proposed against point cloud models \cite{8953645, Wicker2019RobustnessO3, Yang2019AdversarialAA, zheng2019pointcloud, 9010002, 2019arXiv190806062L, 10.1007/978-3-030-58610-2_15, 2019arXiv191211171W, 2020arXiv201100566Z}. Some methods disrupt the global structure by adding \cite{8953645, 2019arXiv190806062L}, removing \cite{zheng2019pointcloud, Wicker2019RobustnessO3}, or shifting \cite{8953645} points. Other methods rely on gradient-based optimization \cite{8803770, 9711142}. For instance, \cite{8803770} adapts 2D approaches such as FGSM and iterative gradient perturbations to point clouds, while \cite{9711142} leverages gradients to generate sparse and almost imperceptible adversarial perturbations.

To counter these attacks, several defense strategies have been developed \cite{2020arXiv201005272W, DBLP:conf/iccv/ZhouCZFZY19, 8803770, 2019arXiv190710764Z, 10.1007/978-3-030-58580-8_20, 9577799}. A very simple yet effective approach is Statistical Outlier Removal (SOR), which mitigates adversarial perturbations by removing outlier points. Furthermore, DUP-Net \cite{DBLP:conf/iccv/ZhouCZFZY19} combines SOR with a PU-Net module that, after point removal, upsamples the point cloud back to its original size. Other approaches use adversarial training (AT) to enhance robustness by explicitly optimizing models on adversarially perturbed point clouds. Methods such as \cite{8803770} and \cite{2019arXiv190710764Z} adapt gradient-based attacks during training, encouraging networks to learn smoother decision boundaries and reducing sensitivity to small geometric perturbations. Finally, data augmentation methods such as PointMixup \cite{10.1007/978-3-030-58580-8_20} and RSMix \cite{9577799} also improve robustness by generating mixed or geometrically transformed samples that regularize the model.

\section{Theoretical Motivation}
\label{sec:theory}
We hypothesize that adversarial fragility in 3D point cloud networks arises from a mismatch between the latent geometry learned by the network and the intrinsic geometry of the underlying object manifold. In this section, we formalize our hypothesis by developing a geometric framework that characterizes robustness in terms of intrinsic–latent alignment. We first describe the ideal condition under which a model faithfully preserves the manifold’s intrinsic structure, then analyze how modern architectures tend to violate this condition, and finally introduce a regularization principle designed to mitigate this problem.

\subsection{Geometric Interpretation of Robustness}
Let $\mathcal{M}$ be an intrinsic surface manifold of a 3D object, characterized by a Riemannian metric $g_{\text{int}}$. The manifold is immersed in $\mathbb{R}^3$ through an embedding $i:\mathcal{M}\!\to\!\mathbb{R}^3$, and a point cloud $X=\{x_i\}_{i=1}^{N}$ represents a finite sampling from it, possibly perturbed by noise or occlusion.

A classifier $f_{\theta}:\mathbb{R}^{N \times 3}\!\to\!\Delta^C$ maps the point cloud $X$ to a probability vector over $C$ object categories. The network simultaneously induces a latent representation space $\mathcal{Z} = f_\theta(\mathcal{M}) \subset \mathbb{R}^d$, where $d$ denotes the dimensionality of the latent feature space.

The geometry of the latent representation space is governed by the Jacobian $Jf_\theta(x)$, which describes how infinitesimal surface perturbations in the input are reflected in latent space. The corresponding local metric is
\begin{equation}
    g_{\text{lat}}(x) = (Jf_\theta(x))^\top Jf_\theta(x).
\end{equation}
For an infinitesimal intrinsic displacement $\delta x \in T_x\mathcal{M}$ (i.e., tangent to the surface), the corresponding change in latent representation satisfies
\begin{equation}
    \|\delta z\|^2 \approx \delta x^\top g_{\text{lat}}(x)\, \delta x.
\end{equation}

Ideally, $g_{\text{lat}}$ should preserve the structure of $g_{\text{int}}$:
\begin{equation}
    g_{\text{lat}}(x) \approx \alpha(x) g_{\text{int}}(x),
    \label{eq:isometry}
\end{equation}
where $\alpha(x)$ is a smooth scalar field allowing uniform local scaling in latent space while preserving the manifold’s intrinsic geometry. 
Here, $g_{\text{int}}$ encodes the true geometry of the surface, where distances are measured as geodesics along the manifold rather than Euclidean distances in $\mathbb{R}^3$. 
When condition~(\ref{eq:isometry}) approximately holds, the mapping behaves locally as a conformal transformation of the intrinsic manifold, meaning that geometry-preserving variations on the surface (e.g., smooth resampling or minor bending) yield proportionally small or negligible changes in latent space, leading to stable predictions.

\subsection{Manifold Misalignment}
In practice, modern point-based networks may deviate significantly from the compatibility condition in Eq.~(\ref{eq:isometry}), leading to distortions of the intrinsic surface geometry in latent space. Perturbations to a 3D shape naturally decompose into intrinsic (on-manifold) and extrinsic (off-manifold) components, and a robust classifier should respond weakly to the former and strongly to the latter. However, architectures such as PointNet, PointNet++, and DGCNN can become highly sensitive to local geometric variations: their Jacobians $Jf_\theta(x)$ may amplify certain intrinsic directions while suppressing others, causing geometry-preserving variations in sampling, curvature, or density to induce disproportionately large shifts in latent space. This effect can be analyzed locally through the geometry induced by $f_\theta$.

Let $\xi_1, \xi_2$ be two nearby points on $\mathcal{M}$ with geodesic intrinsic distance $d_{\text{int}}(\xi_1,\xi_2)$. Under the mapping $f_\theta$, the corresponding latent distance is
\begin{equation}
    d_{\text{lat}}(\xi_1,\xi_2) = \|f_\theta(\xi_1) - f_\theta(\xi_2)\|.
\end{equation}
If $f_\theta$ is locally smooth, the first-order Taylor approximation yields
\begin{equation}
    d_{\text{lat}}(\xi_1,\xi_2)^2 \approx 
    (\xi_2 - \xi_1)^\top (Jf_\theta(\xi_1)^\top Jf_\theta(\xi_1)) (\xi_2 - \xi_1),
\end{equation}
implying that deviations between $g_{\text{int}}$ and $g_{\text{lat}}$ directly modulate the model’s sensitivity to geometric perturbations. 

To quantify this distortion, we define
\begin{equation}
    \mu(x) = \frac{\sigma_{\max}(Jf_\theta(x))}{\sigma_{\min}(Jf_\theta(x))},
\end{equation}
the ratio of the largest and smallest singular values of the Jacobian. Large values of $\mu(x)$ indicate severe anisotropy: the network stretches or compresses the manifold unevenly, amplifying small intrinsic variations into large changes in latent space, potentially increasing vulnerability to adversarial perturbations.

This analysis suggests that robustness can be interpreted as maintaining bounded distortion, i.e., preserving local distances up to a uniform scaling on the manifold. Achieving such bounded distortion requires explicitly constraining the relationship between the intrinsic and latent geometries, which motivates the intrinsic consistency regularization introduced next.

\subsection{Intrinsic Consistency Regularization}
\label{sec:intrinsic_isometry_regularization}
To enforce intrinsic–latent alignment during training, we introduce an intrinsic consistency loss that penalizes discrepancies between changes in the model’s output and the corresponding intrinsic geometric variation. Formally, the loss is defined as
\begin{equation}
\mathcal{L}_{\text{cons}} =
\mathbb{E}_{(X_1, X_2)}
\!\left[
\frac{D(f_\theta(X_1), f_\theta(X_2))}
{\|\Phi(X_1) - \Phi(X_2)\|_F^2 + \epsilon}
\right],
\end{equation}
where $D(\cdot,\cdot)$ measures the divergence between the model outputs for two geometrically close views $(X_1, X_2)$ of the same object, and $\Phi(X)$ denotes a set of intrinsic features encoding local geometric structure (e.g., curvature and diffusion relations) that are approximately invariant under rigid motions. The denominator measures the intrinsic geometric change between the two views, ensuring that the loss penalizes large output differences only when the underlying geometry changes little. This encourages the mapping to maintain local geometric consistency between the intrinsic and latent representations.

Under smoothness assumptions on $f_\theta$ and $\Phi$, both mappings can be
locally linearized around a point $x$:
\begin{equation}
\label{eq:cons_loss}
\begin{aligned}
    f_\theta(X_2) - f_\theta(X_1) &\approx Jf_\theta(x)\,\delta x, \\
    \Phi(X_2) - \Phi(X_1) &\approx J_\Phi(x)\,\delta x,
\end{aligned}
\end{equation}
where $Jf_\theta(x)$ and $J_\Phi(x)$ denote their respective Jacobians with respect to the input coordinates.
Substituting these first-order approximations into Eq.~(\ref{eq:cons_loss}) motivates the approximation
\begin{equation}
    \frac{D(f_\theta(X_1), f_\theta(X_2))}
    {\|\Phi(X_1) - \Phi(X_2)\|_F^2}
    \;\approx\;
    \|Jf_\theta(x)\,J_\Phi^\dagger(x)\|_F^2,
\end{equation}
where $J_\Phi^\dagger(x)$ denotes the pseudoinverse of $J_\Phi(x)$.
Under the ideal compatibility condition, the latent and intrinsic differentials are expected to align locally, motivating the regularized objective
\begin{equation}
\mathcal{L}_{\text{cons}} \approx
\mathbb{E}_x
\!\left[
\big\|Jf_\theta(x)\, J_\Phi^\dagger(x) - I\big\|_F^2
\right],
\end{equation}
where $I$ is the identity matrix.
Minimizing this penalty term encourages $Jf_\theta(x)J_\Phi^\dagger(x)$ to approximate the identity transformation, enforcing directional alignment between the intrinsic and latent differentials and therefore promoting local geometric consistency.
Equivalently, the consistency constraint can be viewed as enforcing a relative Lipschitz bound on $f_\theta$:
\begin{equation}
\frac{\|f_\theta(X_1) - f_\theta(X_2)\|}
{d_{\text{int}}(X_1, X_2)} \le L_{\text{intr}},
\label{eq:relative_lip}
\end{equation}
where $d_{\text{int}}$ denotes the intrinsic distance derived from $\Phi(X)$.
Here, $L_{\text{intr}}$ quantifies the maximal ratio between latent and intrinsic
distances, acting as an intrinsic Lipschitz constant that bounds local distortion
of the manifold geometry.

\vspace{0.3em}
\noindent\textbf{Geometric Implication.}
When the consistency regularizer is sufficiently small, the mapping is encouraged to preserve bounded local distortion with respect to the intrinsic metric. In particular, the singular values of $Jf_\theta(x)J_\Phi^\dagger(x)$ are encouraged to remain bounded within a local neighborhood of the manifold, limiting excessive stretching or compression of intrinsic directions in latent space. Consequently, for sufficiently nearby $x_i,x_j \in \mathcal{M}$, the latent distance can be expected to remain approximately proportional to the intrinsic distance:
\begin{equation}
c_1\, d_{\text{int}}(x_i,x_j)
\;\le\;
\|f_\theta(x_i) - f_\theta(x_j)\|
\;\le\;
c_2\, d_{\text{int}}(x_i,x_j),
\end{equation}
for some positive constants $0 < c_1 < c_2 < \infty$.
Thus, the mapping $f_\theta$ tends to preserve local geometric structure, ensuring that small intrinsic deformations produce proportionate changes in latent space.

\section{Method}
\label{sec:method}

\subsection{Manifold-Aligned Point Recognition (MAPR)}
Given a point cloud $X=\{x_i\}_{i=1}^{N}\!\subset\!\mathbb{R}^3$, we compute a set of intrinsic features $\Phi(X)\!\in\!\mathbb{R}^{N\times C_{\text{intr}}}$ that capture the manifold’s intrinsic geometry, including curvature and multi-scale diffusion structure. These intrinsic features are then concatenated with the raw coordinates to form the augmented point representation $\widetilde{X}=[X~|~\Phi(X)]\!\in\!\mathbb{R}^{N\times(3+C_{\text{intr}})}$. In addition, MAPR incorporates an intrinsic consistency regularization term that encourages stable predictions under small geometry-preserving perturbations. This regularizer constrains the latent representation to vary consistently with the intrinsic structure of the underlying surface. The full training process is summarized in Algorithm~\ref{alg:mapr}.

Importantly, MAPR is deliberately constructed from a small set of components directly derived from the proposed geometric interpretation of robustness. The goal is not to maximize performance through the combination of multiple robustness heuristics, but to study the extent to which intrinsic–latent alignment alone can improve robustness across diverse architectures and perturbations.

\paragraph{Intrinsic features.}
The intrinsic feature map $\Phi(X)$ comprises two complementary components:

\begin{itemize}
\item \textit{Curvature estimates.}
We estimate local curvature by applying the random-walk Laplacian $L_{\mathrm{rw}}$ to the point coordinates. For each point $x_i$, the quantity
\[
\kappa(x_i)=\|(L_{\mathrm{rw}} X)_i\|_2
\]
provides a differential estimate of local curvature, reflecting how far the point deviates from its neighbors.

\item \textit{Multi-scale diffusion descriptors.}
We construct a $k$-NN graph and model multi-scale diffusion through powers of its row-stochastic adjacency matrix $A$. For each point, the features $A^t X$ and $A^t \mathbf{1}$ at diffusion steps
\[
t\in\{1,2,4,8\}
\]
encode how local geometry and sampling density evolve across increasingly larger neighborhoods, providing a coarse-to-fine intrinsic descriptor.
\end{itemize}

Additional implementation details of the intrinsic feature construction are provided in Appendix A.

\paragraph{Intrinsic consistency loss.}
Following the intrinsic consistency principle introduced in Sec.~\ref{sec:intrinsic_isometry_regularization}, we define the following intrinsic consistency loss.

Each input point cloud $X$ is first perturbed to obtain
\[
X'=\texttt{Perturb}(X),
\]
where a small random rotation and Gaussian jitter are applied to the point coordinates. Intrinsic features are then computed for both views to obtain the augmented inputs
\[
\widetilde{X}=[X~|~\Phi(X)],
\qquad
\widetilde{X'}=[X'~|~\Phi(X')].
\]

The corresponding logits are given by
\[
\ell=f_\theta(\widetilde{X}),
\qquad
\ell'=f_\theta(\widetilde{X'}).
\]

Applying the softmax operator to the logits yields
\[
p=\mathrm{softmax}(\ell),
\qquad
q=\mathrm{softmax}(\ell'),
\]
which denote the predicted class distributions for the original and perturbed inputs, respectively. The consistency loss is defined as
\begin{equation}
\mathcal{L}_{\text{cons}}
=
\frac{1}{B}\sum_{i=1}^{B}
\frac{
D_{\mathrm{SKL}}\!\left(p^{(i)},q^{(i)}\right)
}
{
\|\Phi(X^{(i)})-\Phi({X'}^{(i)})\|_F^2+\epsilon
},
\end{equation}
where $B$ is the batch size, $\epsilon$ ensures numerical stability, and
$D_{\mathrm{SKL}}$ denotes the symmetric KL divergence:
\begin{equation}
D_{\mathrm{SKL}}(p,q)
=
\frac{1}{2}
\left[
D_{\mathrm{KL}}(p \Vert q)
+
D_{\mathrm{KL}}(q \Vert p)
\right].
\end{equation}

This term penalizes large changes in the predicted distributions under geometry-preserving perturbations, encouraging the latent representations to remain locally aligned with the intrinsic structure of the underlying surface.

\paragraph{Training loss.}
The overall loss combines a standard classification loss with the proposed intrinsic consistency regularization:
\begin{equation}
\label{eq:total_loss}
\mathcal{L}
=
\mathcal{L}_{\text{cls}}
+
\lambda_{\text{lip}}\mathcal{L}_{\text{cons}},
\end{equation}
where $\mathcal{L}_{\text{cls}}$ is the standard cross-entropy loss and
$\lambda_{\text{lip}}$ controls the strength of the regularizer.

By discouraging disproportionate changes in prediction space under geometry-preserving perturbations, the model learns latent representations that are better aligned with the intrinsic geometry of the underlying surface.

\begin{algorithm}[t]
\caption{Manifold-Aligned Point Recognition (MAPR)}
\label{alg:mapr}
\begin{algorithmic}[1]
\Require Dataset $\mathcal{Q}$, labels $Y$, backbone $f_\theta$, intrinsic map $\Phi$
\For{each training epoch}
    \ForAll{batch $(X, Y) \sim \mathcal{Q}$}
        \State Generate perturbed view $X' = \texttt{Perturb}(X)$
        \State Compute intrinsic features $\Phi(X)$, $\Phi(X')$
        \State Construct augmented inputs
        \Statex \hspace{\algorithmicindent} \hspace{\algorithmicindent} \hspace{\algorithmicindent}
        \hspace{\algorithmicindent} $\widetilde{X} = \operatorname{Concat}(X, \Phi(X))$,
        \Statex \hspace{\algorithmicindent} \hspace{\algorithmicindent}
        \hspace{\algorithmicindent}
        \hspace{\algorithmicindent}$\widetilde{X'} = \operatorname{Concat}(X', \Phi(X'))$
        \State Compute logits $\ell = f_\theta(\widetilde{X})$, $\ell' = f_\theta(\widetilde{X'})$
        \State Compute classification loss $\mathcal{L}_{\text{cls}}$
        \State Compute consistency loss $\mathcal{L}_{\text{cons}}$
        \State Update $\theta$ using 
        \Statex \hspace{\algorithmicindent}
        \hspace{\algorithmicindent}
        \hspace{\algorithmicindent}
        \hspace{\algorithmicindent}
        $\mathcal{L} = \mathcal{L}_{\text{cls}} + \lambda_{\text{lip}}\,\mathcal{L}_{\text{cons}}$
    \EndFor
\EndFor
\State \Return trained model $f_\theta$
\end{algorithmic}
\end{algorithm}

\section{Experiments}
\label{sec:experiments}

\begin{table*}[h]
\caption{Clean and robust accuracies (\%) on the ModelNet40 dataset for five representative 3D backbones under Vanilla, AT, and MAPR training, evaluated under multiple adversarial perturbations. Results are reported without defense (top block) and with the SOR defense (bottom block). Best results within each block are highlighted in bold. The final column reports the average robustness accuracy (excluding the clean accuracy).}
\centering
\resizebox{\textwidth}{!}{
\begin{tabular}{llccccccccccc}
\toprule
\textbf{Model} & \textbf{Method} & \textbf{Defense} & \textbf{Clean} 
& \textbf{SMA-100} & \textbf{PGD-20 $\ell_2$} & \textbf{PGD-20 $\ell_\infty$} 
& \textbf{FGSM} & \textbf{BIM} & \textbf{Add-100} & \textbf{T-PGD} & \textbf{SI-PGD} & \textbf{Avg} \\
\midrule

\multirow{6}{*}{PointNet}
& Vanilla & $-$ 
& \textbf{89.95} & 87.88 & 80.02 & 11.35 & 62.84 & 14.42 & 11.59 & 2.35 & 32.74 & 37.90 \\
& AT & $-$ 
& 89.06 & \textbf{88.21} & \textbf{83.14} & 21.56 & 70.38 & 24.68 & 16.29 & 7.50 & 41.94 & 44.21 \\
& MAPR & $-$ 
& 89.83 & 87.76 & 82.94 & \textbf{32.82} & \textbf{70.95} & \textbf{37.16} & \textbf{36.63} & \textbf{23.10} & \textbf{53.36} & \textbf{53.09} \\

\cmidrule(lr){2-13}

& Vanilla & SOR 
& \textbf{89.91} & 87.32 & 85.01 & 42.91 & 70.79 & 45.66 & \textbf{81.32} & 43.76 & 51.30 & 63.51 \\
& AT & SOR 
& 88.94 & \textbf{87.64} & 85.66 & 51.01 & 76.30 & 54.54 & 78.81 & \textbf{51.05} & 60.29 & 68.16 \\
& MAPR & SOR 
& 89.06 & 87.16 & \textbf{87.40} & \textbf{59.97} & \textbf{77.84} & \textbf{62.56} & 70.10 & 45.30 & \textbf{67.75} & \textbf{69.76} \\

\midrule

\multirow{6}{*}{PointNet++}
& Vanilla & $-$ 
& 91.25 & 89.38 & 84.36 & 11.55 & 73.46 & 18.19 & 16.49 & 9.28 & 38.01 & 42.59 \\
& AT & $-$ 
& \textbf{91.98} & \textbf{90.32} & \textbf{88.49} & 19.49 & 78.44 & 29.54 & 22.57 & 11.87 & 51.70 & 49.05 \\
& MAPR & $-$ 
& 90.15 & 89.10 & 87.68 & \textbf{61.79} & \textbf{81.08} & \textbf{64.42} & \textbf{56.08} & \textbf{47.53} & \textbf{71.39} & \textbf{69.88} \\

\cmidrule(lr){2-13}

& Vanilla & SOR 
& 89.91 & 88.17 & 87.52 & 45.46 & 72.29 & 51.18 & 79.13 & 36.91 & 63.29 & 65.49 \\
& AT & SOR 
& \textbf{90.52} & \textbf{89.83} & \textbf{89.30} & 53.97 & 74.11 & 58.06 & \textbf{83.91} & 39.99 & 68.56 & 69.72 \\
& MAPR & SOR 
& 89.59 & 86.63 & 89.14 & \textbf{75.45} & \textbf{82.41} & \textbf{77.43} & 79.42 & \textbf{60.94} & \textbf{80.51} & \textbf{78.99} \\

\midrule

\multirow{6}{*}{DGCNN}
& Vanilla & $-$
& 91.41 & 89.22 & 55.27 & 7.50 & 80.88 & 23.18 & 25.69 & 39.22 & 43.96 & 45.62 \\
& AT & $-$ 
& \textbf{91.57} & \textbf{89.83} & 85.70 & 42.63 & \textbf{84.68} & 57.54 & \textbf{45.30} & \textbf{50.89} & \textbf{71.07} & \textbf{65.95} \\
& MAPR & $-$
& 91.41 & 88.01 & \textbf{86.18} & \textbf{50.20} & 81.32 & \textbf{59.36} & 43.64 & 47.16 & 69.49 & 65.67 \\

\cmidrule(lr){2-13}

& Vanilla & SOR
& \textbf{88.70} & 84.28 & 84.97 & 49.11 & 75.08 & 57.09 & 70.42 & 55.75 & 66.00 & 67.84 \\
& AT & SOR 
& 88.53 & \textbf{84.48} & \textbf{87.93} & 65.88 & \textbf{80.51} & \textbf{71.11} & \textbf{83.31} & \textbf{63.86} & \textbf{75.85} & \textbf{76.62} \\
& MAPR & SOR
& 88.41 & 81.32 & 87.44 & \textbf{69.53} & 77.47 & 70.50 & 74.76 & 60.98 & 74.76 & 74.59 \\

\midrule

\multirow{6}{*}{CurveNet}
& Vanilla & $-$ 
& 90.32 & 89.30 & 85.01 & 63.90 & 84.72 & 69.45 & 0.41 & 60.78 & 76.13 & 66.21 \\
& AT & $-$ 
& 90.64 & \textbf{90.88} & \textbf{88.86} & \textbf{75.61} & \textbf{86.99} & \textbf{79.17} & 2.23 & \textbf{66.49} & \textbf{81.52} & 71.47 \\
& MAPR & $-$ 
& \textbf{91.21} & 89.47 & 87.60 & 69.45 & 85.98 & 73.34 & \textbf{42.34} & 66.00 & 78.48 & \textbf{74.08} \\

\cmidrule(lr){2-13}

& Vanilla & SOR 
& 89.30 & 88.29 & 88.49 & 76.22 & 83.43 & 78.04 & 43.15 & 74.15 & 81.16 & 76.62 \\
& AT & SOR 
& 90.07 & \textbf{89.95} & 89.95 & \textbf{81.48} & 86.83 & \textbf{82.33} & 42.63 & \textbf{77.51} & \textbf{84.32} & 79.38 \\
& MAPR & SOR 
& \textbf{90.68} & 89.10 & \textbf{90.68} & 78.40 & \textbf{87.20} & 80.15 & \textbf{75.57} & 76.74 & 83.71 & \textbf{82.69} \\

\midrule

\multirow{6}{*}{PointMLP}
& Vanilla & $-$ 
& 89.67 & 88.86 & 69.12 & 10.78 & 77.11 & 24.35 & 6.28 & 21.60 & 45.10 & 42.90 \\
& AT & $-$ 
& \textbf{91.45} & \textbf{90.36} & 85.29 & 32.78 & \textbf{83.95} & 47.89 & 10.21 & 36.95 & 65.36 & 56.60 \\
& MAPR & $-$ 
& 90.96 & 88.65 & \textbf{88.49} & \textbf{55.83} & 82.25 & \textbf{60.66} & \textbf{84.44} & \textbf{50.89} & \textbf{69.45} & \textbf{72.58} \\

\cmidrule(lr){2-13}

& Vanilla & SOR 
& 87.40 & 86.22 & 85.13 & 45.42 & 69.00 & 51.46 & 74.39 & 41.86 & 63.45 & 64.62 \\
& AT & SOR 
& \textbf{90.56} & \textbf{89.83} & 89.42 & 59.44 & 79.82 & 65.24 & 81.44 & 53.97 & 74.55 & 74.21 \\
& MAPR & SOR 
& 90.07 & 86.55 & \textbf{89.91} & \textbf{70.02} & \textbf{81.24} & \textbf{71.07} & \textbf{90.19} & \textbf{58.39} & \textbf{77.59} & \textbf{78.12} \\

\bottomrule
\end{tabular}
}
\label{tab:all_backbones_results_MN40}
\end{table*}

\begin{table*}[h]
\caption{Clean and robust accuracies (\%) on the ScanObjectNN dataset for five representative 3D backbones under Vanilla, AT, and MAPR training, evaluated under multiple adversarial perturbations. Results are reported without defense (top block) and with the SOR defense (bottom block). Best results within each block are highlighted in bold. The final column reports the average robustness accuracy (excluding the clean accuracy).}
\centering
\resizebox{\textwidth}{!}{
\begin{tabular}{llccccccccccc}
\toprule
\textbf{Model} & \textbf{Method} & \textbf{Defense} & \textbf{Clean} 
& \textbf{SMA-100} & \textbf{PGD-20 $\ell_2$} & \textbf{PGD-20 $\ell_\infty$} 
& \textbf{FGSM} & \textbf{BIM} & \textbf{Add-100} & \textbf{T-PGD} & \textbf{SI-PGD} & \textbf{Avg} \\
\midrule

\multirow{6}{*}{PointNet}
& Vanilla & $-$ 
& 67.77 & 65.06 & 40.22 & 1.53 & 26.13 & 2.05 & 0.76 & 0.28 & 5.62 & 17.71 \\
& AT & $-$ 
& 67.87 & 66.52 & 47.74 & 3.16 & \textbf{33.38} & 4.02 & 1.49 & 0.62 & 9.02 & 20.74 \\
& MAPR & $-$ 
& \textbf{69.64} & \textbf{68.22} & \textbf{59.75} & \textbf{6.00} & 26.75 & \textbf{7.43} & \textbf{1.63} & \textbf{1.53} & \textbf{11.10} & \textbf{22.80} \\

\cmidrule(lr){2-13}

& Vanilla & SOR 
& 67.04 & 64.02 & 53.40 & 14.75 & 42.19 & 17.56 & 40.39 & 20.44 & 17.28 & 33.75 \\
& AT & SOR 
& 67.28 & 65.48 & 57.74 & \textbf{21.96} & \textbf{45.94} & \textbf{25.09} & \textbf{43.23} & \textbf{27.20} & \textbf{22.97} & \textbf{38.70} \\
& MAPR & SOR 
& \textbf{69.15} & \textbf{67.49} & \textbf{63.88} & 12.98 & 27.34 & 14.23 & 43.03 & 2.85 & 18.29 & 31.26 \\

\midrule

\multirow{6}{*}{PointNet++}
& Vanilla & $-$ 
& \textbf{79.46} & 77.79 & 56.28 & 0.00 & 32.27 & 0.35 & 13.22 & 2.67 & 3.96 & 23.32 \\
& AT & $-$ 
& 79.08 & \textbf{78.21} & \textbf{66.52} & 0.59 & 37.65 & 1.70 & 24.57 & 4.20 & 10.83 & 28.03 \\
& MAPR & $-$ 
& 74.57 & 72.14 & 65.75 & \textbf{16.10} & \textbf{48.61} & \textbf{19.08} & \textbf{33.48} & \textbf{13.95} & \textbf{30.71} & \textbf{37.48} \\

\cmidrule(lr){2-13}

& Vanilla & SOR 
& \textbf{79.04} & 76.93 & 66.59 & 6.63 & 42.61 & 9.58 & 53.33 & 8.08 & 12.11 & 34.48 \\
& AT & SOR 
& 78.42 & \textbf{77.65} & \textbf{71.41} & 14.78 & 47.15 & 18.70 & \textbf{62.80} & 11.97 & 22.28 & 40.84 \\
& MAPR & SOR 
& 72.83 & 71.27 & 68.01 & \textbf{27.86} & \textbf{51.32} & \textbf{30.22} & 53.02 & \textbf{18.08} & \textbf{38.86} & \textbf{44.83} \\

\midrule

\multirow{6}{*}{DGCNN}
& Vanilla & $-$ 
& 76.75 & 74.88 & 22.80 & 0.52 & 45.77 & 5.59 & \textbf{32.13} & 16.00 & 11.73 & 26.18 \\
& AT & $-$ 
& 75.75 & 74.77 & 57.08 & 6.21 & \textbf{53.37} & \textbf{13.67} & 27.38 & \textbf{18.88} & \textbf{24.81} & \textbf{34.52} \\
& MAPR & $-$ 
& \textbf{78.35} & \textbf{76.75} & \textbf{63.36} & \textbf{7.18} & 51.01 & 12.63 & 7.49 & 16.31 & 23.80 & 32.32 \\

\cmidrule(lr){2-13}

& Vanilla & SOR 
& 72.62 & 70.06 & 53.82 & 9.65 & 39.00 & 14.47 & \textbf{64.75} & 19.15 & 22.48 & 36.67 \\
& AT & SOR 
& 71.72 & 68.98 & 65.89 & \textbf{19.54} & \textbf{47.85} & \textbf{26.41} & 58.05 & \textbf{22.28} & \textbf{35.95} & \textbf{43.12} \\
& MAPR & SOR 
& \textbf{73.66} & \textbf{71.37} & \textbf{68.60} & 18.53 & 46.50 & 23.18 & 54.48 & 17.04 & 31.99 & 41.46 \\

\midrule

\multirow{6}{*}{CurveNet}
& Vanilla & $-$ 
& 76.27 & 75.47 & 51.73 & 14.64 & 54.51 & 25.29 & \textbf{5.66} & \textbf{23.49} & 36.26 & 35.88 \\
& AT & $-$ 
& 77.10 & 75.61 & 69.81 & \textbf{27.72} & 57.98 & \textbf{34.07} & 2.67 & 22.10 & 44.07 & 41.75 \\
& MAPR & $-$ 
& \textbf{80.12} & \textbf{78.17} & \textbf{72.24} & 26.34 & \textbf{63.67} & 32.93 & 2.91 & 20.58 & \textbf{44.24} & \textbf{42.64} \\

\cmidrule(lr){2-13}

& Vanilla & SOR 
& 74.53 & 72.41 & 68.32 & \textbf{37.23} & 52.74 & \textbf{42.78} & 40.01 & \textbf{34.21} & 47.68 & 49.42 \\
& AT & SOR 
& 75.78 & 74.77 & 73.70 & 35.77 & 56.77 & 40.01 & 27.90 & 25.99 & \textbf{49.06} & 48.00 \\
& MAPR & SOR 
& \textbf{79.15} & \textbf{76.82} & \textbf{75.85} & 33.31 & \textbf{58.74} & 38.20 & \textbf{52.29} & 20.96 & 47.99 & \textbf{50.52} \\

\midrule

\multirow{6}{*}{PointMLP}
& Vanilla & $-$ 
& 78.87 & 77.97 & 25.57 & 0.35 & 51.28 & 2.95 & 24.08 & \textbf{18.32} & 12.42 & 26.62 \\
& AT & $-$ 
& \textbf{83.66} & \textbf{82.44} & 65.09 & 4.37 & \textbf{55.41} & 8.81 & 11.17 & 15.79 & \textbf{21.17} & 33.03 \\
& MAPR & $-$ 
& 83.34 & 82.06 & \textbf{73.14} & \textbf{6.18} & 50.73 & \textbf{8.88} & \textbf{61.42} & 8.19 & 18.22 & \textbf{38.60} \\

\cmidrule(lr){2-13}

& Vanilla & SOR 
& 74.60 & 74.15 & 59.47 & 9.20 & 44.97 & 18.36 & 67.04 & \textbf{25.78} & 28.28 & 40.91 \\
& AT & SOR 
& \textbf{81.37} & \textbf{79.56} & 75.02 & \textbf{12.21} & \textbf{51.49} & \textbf{18.91} & 55.97 & 18.98 & \textbf{32.13} & \textbf{43.03} \\
& MAPR & SOR 
& 81.33 & 79.29 & \textbf{75.54} & 9.85 & 47.81 & 11.87 & \textbf{78.07} & 7.43 & 22.28 & 41.52 \\

\bottomrule
\end{tabular}
}
\label{tab:all_backbones_results_ScanObjectNN}
\end{table*}

\subsection{Experimental Setup}
\label{sec:experimental_setup}
We evaluate the robustness and generalization capabilities of MAPR across multiple architectures, datasets, and adversarial attack settings.

\paragraph{Models and Datasets.}
We evaluate robustness across five representative point cloud classification architectures: PointNet \citep{Qi_2017_CVPR}, PointNet++ \citep{NIPS2017_d8bf84be}, DGCNN \citep{10.1145/3326362}, CurveNet \citep{curvenet2021}, and PointMLP \citep{Ma2022RethinkingND}. For each architecture, we compare three training settings: the vanilla model, adversarial training (AT), and MAPR.

We conduct experiments on two datasets: ModelNet40 \citep{Wu_2015_CVPR} and ScanObjectNN \citep{9009007}. ModelNet40 is a synthetic CAD dataset containing 12,311 shapes across 40 object categories. We follow the standard split of 9,843 shapes for training and 2,468 for testing. Each shape is uniformly sampled to 2,048 points and normalized to a unit sphere.

To evaluate generalization beyond clean synthetic CAD models, we further evaluate on ScanObjectNN, a more realistic and challenging dataset composed of scanned real-world objects with background clutter, occlusion, and acquisition noise. We use its hardest split, PB-T50-RS. As with ModelNet40, point clouds are sampled to 2,048 points and normalized.

\paragraph{Attacks.}
\label{sec:attacks}
Each model is evaluated under several adversarial attacks. In addition to standard attacks (SMA drop, PGD-$\ell_{2}$, PGD-$\ell_{\infty}$, FGSM, BIM), we include three additional attacks (Add-Points, T-PGD, SI-PGD) as complementary evaluations for a more complete robustness assessment.
\begin{itemize}
    \item \textbf{SMA drop}: The SMA drop attack \cite{zheng2019pointcloud} removes the points with the largest saliency scores.
    \item \textbf{PGD-$\ell_{2}$}: Projected Gradient Descent (PGD) $\ell_2$ \cite{Madry2017} applies projected gradient ascent under an $\ell_2$-bounded perturbation.
    \item \textbf{PGD-$\ell_{\infty}$}: Projected Gradient Descent (PGD) $\ell_{\infty}$ \cite{Madry2017} applies projected gradient ascent under an $\ell_{\infty}$-bounded perturbation.
    \item \textbf{FGSM}: Fast Gradient Sign Method (FGSM) \cite{goodfellow2014explaining} generates a single-step perturbation in the gradient sign direction under an $\ell_\infty$ bound.
    \item \textbf{BIM}: Basic Iterative Method (BIM) \cite{2016arXiv160702533K} extends FGSM into an iterative procedure, repeatedly applying small gradient-sign steps under an $\ell_\infty$ bound.
    \item \textbf{Add-$k$}: We optimize the coordinates of $k$ added points through gradient ascent and append them to the original cloud to form the adversarial example.
    \item \textbf{T-PGD}: Transfer PGD (T-PGD) generates adversarial examples using an ensemble-based PGD attack that averages gradients from multiple surrogate models and updates them with momentum to improve transferability.
    \item \textbf{SI-PGD}: Shape-Invariant PGD (SI-PGD) augments PGD with a penalty that constrains changes in local $k$-NN geometry, encouraging perturbations that preserve the underlying shape structure.
\end{itemize}
For Add-$k$ and SMA drop, we add/drop 100 points (SMA-100, Add-100). For PGD-$\ell_{2}$ and PGD-$\ell_{\infty}$, we use 20 iterations (PGD-20 $\ell_{2}$, PGD-20 $\ell_{\infty}$). The same attack protocol is used for both ModelNet40 and ScanObjectNN.

\paragraph{SOR defense.}
In addition to the standard evaluation protocol, we also report results with Statistical Outlier Removal (SOR) defense \cite{DBLP:conf/iccv/ZhouCZFZY19}. SOR removes points whose average distance to their $k$ nearest neighbors exceeds a threshold determined by the global point cloud statistics. Following prior work, we apply SOR as a preprocessing step before classification using $k=2$ nearest neighbors and a standard deviation threshold of $\alpha=1.1$.

\paragraph{Training details.}
All models are trained for 100 epochs using the Adam optimizer \citep{kingma2014adam}. For each architecture, we perform a grid search over multiple learning rates and select the rate that achieves the highest classification accuracy for the vanilla model. Selected learning rates are: $0.002$ (PointNet), $0.0005$ (PointNet++), $0.003$ (DGCNN), $0.002$ (CurveNet), and $0.0003$ (PointMLP), for ModelNet40. For ScanObjectNN, the selected learning rates are: $0.002$ (PointNet), $0.0005$ (PointNet++), $0.001$ (DGCNN), $0.002$ (CurveNet), and $0.002$ (PointMLP). The same learning rates are used for the AT and MAPR variants to ensure fair comparisons. During training, learning rates decay by a factor of 0.7 every 20 epochs. We evaluate all models using the final checkpoint at epoch 100. The batch size is fixed to 12 across all experiments, determined by the memory constraints of the most demanding configuration (MAPR-PointMLP).

The total loss is defined in Eq.~\ref{eq:total_loss}, where $\lambda_{\text{lip}}=0$ for vanilla and AT models. For MAPR, $\lambda_{\text{lip}}$ is linearly increased from 0 to $\lambda_{\max}$ during the first 15 epochs and kept constant thereafter. We evaluate each MAPR model under several values of $\lambda_{\text{lip}}$ and select the value with the highest clean accuracy as $\lambda_{\max}$ (see Appendix B). The classification term uses standard cross-entropy loss. The intrinsic consistency term is computed using the symmetric KL divergence between the softmax distributions of the original and perturbed point clouds. This quantity is normalized by the intrinsic feature variation between the original and perturbed inputs, denoted by $\Delta_{\text{intr}}+\epsilon$.

For comparison, we also evaluate PGD-based adversarial training (AT) with an $\ell_2$-bounded attack. For each training batch, adversarial point clouds are generated using 20 PGD iterations with $\epsilon=0.05$ and step size $0.005$. The model is then optimized on both clean and adversarial point clouds. The final loss is defined as
\[
\mathcal{L}_{\mathrm{AT}}
=
\alpha \mathcal{L}_{\mathrm{CE}}(f_\theta(X), y)
+
(1-\alpha)\mathcal{L}_{\mathrm{CE}}(f_\theta(X_{\mathrm{adv}}), y),
\]
where $\alpha=0.5$, giving equal weight to the clean and adversarial losses.

\begin{table*}[h]
\caption{Ablation study evaluating the contribution of each MAPR component across five representative 3D backbones on the ModelNet40 dataset. We report clean and robust accuracies (\%) for models trained with intrinsic features only, Lipschitz regularization only, and the full MAPR formulation. Best results within each backbone are highlighted in bold. The final column reports the average robustness accuracy (excluding the clean accuracy).}
\centering
\resizebox{\textwidth}{!}{
\begin{tabular}{llccccccccccc}
\toprule
\textbf{Model} & \textbf{Method} & \textbf{Defense} & \textbf{Clean} 
& \textbf{SMA-100} & \textbf{PGD-20 $\ell_2$} & \textbf{PGD-20 $\ell_\infty$} 
& \textbf{FGSM} & \textbf{BIM} & \textbf{Add-100} & \textbf{T-PGD} & \textbf{SI-PGD} & \textbf{Avg} \\
\midrule

\multirow{3}{*}{PointNet}
& Intrinsic-only & $-$ 
& 89.47 & 87.40 & 78.85 & 15.92 & 65.96 & 18.88 & 7.41 & 6.20 & 32.33 & 39.12 \\
& Lip-only & $-$ 
& 89.38 & 87.44 & 80.19 & 16.94 & 64.63 & 20.26 & 24.31 & 6.85 & 39.10 & 42.47 \\
& Full MAPR & $-$ 
& \textbf{89.83} & \textbf{87.76} & \textbf{82.94} & \textbf{32.82} & \textbf{70.95} & \textbf{37.16} & \textbf{36.63} & \textbf{23.10} & \textbf{53.36} & \textbf{53.09} \\

\midrule

\multirow{3}{*}{PointNet++}
& Intrinsic-only & $-$ 
& 90.92 & 90.56 & 85.21 & 8.59 & 73.82 & 16.82 & 25.12 & 10.49 & 36.10 & 43.34 \\
& Lip-only & $-$ 
& \textbf{92.14} & \textbf{90.92} & 86.18 & 11.47 & 75.81 & 19.12 & 50.49 & 18.03 & 54.09 & 50.76 \\
& Full MAPR & $-$ 
& 90.15 & 89.10 & \textbf{87.68} & \textbf{61.79} & \textbf{81.08} & \textbf{64.42} & \textbf{56.08} & \textbf{47.53} & \textbf{71.39} & \textbf{69.88} \\

\midrule

\multirow{3}{*}{DGCNN}
& Intrinsic-only & $-$ 
& 90.40 & 88.17 & 55.19 & 7.90 & 78.00 & 22.93 & 20.75 & 36.43 & 43.48 & 44.11 \\
& Lip-only & $-$ 
& 91.09 & \textbf{89.71} & 67.26 & 14.26 & \textbf{83.35} & 33.95 & 30.43 & \textbf{49.55} & 52.76 & 52.66 \\
& Full MAPR & $-$ 
& \textbf{91.41} & 88.01 & \textbf{86.18} & \textbf{50.20} & 81.32 & \textbf{59.36} & \textbf{43.64} & 47.16 & \textbf{69.49} & \textbf{65.67} \\

\midrule

\multirow{3}{*}{CurveNet}
& Intrinsic-only & $-$ 
& 89.30 & 89.99 & 84.28 & 62.97 & 84.68 & 69.33 & 2.63 & 60.17 & 76.22 & 66.28 \\
& Lip-only & $-$ 
& 90.88 & \textbf{90.15} & 84.93 & 65.80 & \textbf{86.10} & 72.49 & 32.70 & 65.11 & \textbf{78.69} & 72.00 \\
& Full MAPR & $-$ 
& \textbf{91.21} & 89.47 & \textbf{87.60} & \textbf{69.45} & 85.98 & \textbf{73.34} & \textbf{42.34} & \textbf{66.00} & 78.48 & \textbf{74.08} \\

\midrule

\multirow{3}{*}{PointMLP}
& Intrinsic-only & $-$ 
& 89.38 & 89.18 & 68.68 & 8.18 & 78.69 & 22.24 & 11.06 & 19.69 & 43.84 & 42.70 \\
& Lip-only & $-$ 
& \textbf{91.13} & \textbf{90.28} & 81.40 & 4.62 & 78.28 & 13.13 & 76.13 & 25.61 & 41.37 & 51.35 \\
& Full MAPR & $-$ 
& 90.96 & 88.65 & \textbf{88.49} & \textbf{55.83} & \textbf{82.25} & \textbf{60.66} & \textbf{84.44} & \textbf{50.89} & \textbf{69.45} & \textbf{72.58} \\

\bottomrule
\end{tabular}
}
\label{tab:ablation_mapr_components_MN40}
\end{table*}

\begin{table*}[h]
\caption{Ablation study evaluating the contribution of each MAPR component across five representative 3D backbones on the ScanObjectNN dataset. We report clean and robust accuracies (\%) for models trained with intrinsic features only, Lipschitz regularization only, and the full MAPR formulation. Best results within each backbone are highlighted in bold. The final column reports the average robustness accuracy (excluding the clean accuracy).}
\centering
\resizebox{\textwidth}{!}{
\begin{tabular}{llccccccccccc}
\toprule
\textbf{Model} & \textbf{Method} & \textbf{Defense} & \textbf{Clean} 
& \textbf{SMA-100} & \textbf{PGD-20 $\ell_2$} & \textbf{PGD-20 $\ell_\infty$} 
& \textbf{FGSM} & \textbf{BIM} & \textbf{Add-100} & \textbf{T-PGD} & \textbf{SI-PGD} & \textbf{Avg} \\
\midrule

\multirow{3}{*}{PointNet}
& Intrinsic-only & $-$ 
& 66.41 & 64.99 & 39.07 & 1.35 & 26.41 & 1.94 & 0.66 & 0.10 & 5.76 & 17.54 \\
& Lip-only & $-$ 
& 68.22 & 67.04 & 42.16 & 0.97 & \textbf{30.78} & 1.77 & 0.73 & 0.21 & 5.86 & 18.19 \\
& Full MAPR & $-$ 
& \textbf{69.64} & \textbf{68.22} & \textbf{59.75} & \textbf{6.00} & 26.75 & \textbf{7.43} & \textbf{1.63} & \textbf{1.53} & \textbf{11.10} & \textbf{22.80} \\

\midrule

\multirow{3}{*}{PointNet++}
& Intrinsic-only & $-$ 
& \textbf{79.42} & \textbf{78.11} & 55.83 & 0.03 & 32.65 & 0.42 & 13.57 & 2.46 & 3.64 & 23.34 \\
& Lip-only & $-$ 
& \textbf{79.42} & 77.86 & 57.63 & 0.03 & 36.36 & 0.24 & 23.28 & 2.32 & 8.85 & 25.32 \\
& Full MAPR & $-$ 
& 74.57 & 72.14 & \textbf{65.75} & \textbf{16.10} & \textbf{48.61} & \textbf{19.08} & \textbf{33.48} & \textbf{13.95} & \textbf{30.71} & \textbf{37.48} \\

\midrule

\multirow{3}{*}{DGCNN}
& Intrinsic-only & $-$ 
& 76.75 & 74.88 & 22.80 & 0.52 & 45.77 & 5.59 & 32.13 & 16.00 & 11.73 & 26.18 \\
& Lip-only & $-$ 
& 76.93 & 75.75 & 32.93 & 1.18 & \textbf{56.70} & 7.70 & \textbf{36.88} & \textbf{25.47} & 17.97 & 31.82 \\
& Full MAPR & $-$ 
& \textbf{78.35} & \textbf{76.75} & \textbf{63.36} & \textbf{7.18} & 51.01 & \textbf{12.63} & 7.49 & 16.31 & \textbf{23.80} & \textbf{32.32} \\

\midrule

\multirow{3}{*}{CurveNet}
& Intrinsic-only & $-$ 
& 76.27 & 76.20 & 51.46 & 15.16 & 54.20 & 24.46 & 5.66 & \textbf{23.84} & 36.33 & 35.41 \\
& Lip-only & $-$ 
& 78.31 & 77.17 & 64.68 & 14.85 & 63.43 & 22.76 & \textbf{16.45} & 20.89 & 38.48 & 39.84 \\
& Full MAPR & $-$ 
& \textbf{80.12} & \textbf{78.17} & \textbf{72.24} & \textbf{26.34} & 63.67 & \textbf{32.93} & 2.91 & 20.58 & \textbf{44.24} & \textbf{42.64} \\

\midrule

\multirow{3}{*}{PointMLP}
& Intrinsic-only & $-$ 
& 79.22 & 77.41 & 23.25 & 0.03 & 48.68 & 1.73 & 12.73 & \textbf{14.54} & 8.19 & 23.32 \\
& Lip-only & $-$ 
& 83.10 & \textbf{83.52} & 48.75 & 0.24 & \textbf{60.83} & 1.08 & 54.34 & 13.43 & 8.78 & 33.87 \\
& Full MAPR & $-$ 
& \textbf{83.34} & 82.06 & \textbf{73.14} & \textbf{6.18} & 50.73 & \textbf{8.88} & \textbf{61.42} & 8.19 & \textbf{18.22} & \textbf{38.60} \\

\bottomrule
\end{tabular}
}
\label{tab:ablation_mapr_components_SObjNN}
\end{table*}

\subsection{Main Results}
\label{sec:main_results}

We evaluate five representative architectures under multiple adversarial perturbations on both ModelNet40 and ScanObjectNN. Tables~\ref{tab:all_backbones_results_MN40} and~\ref{tab:all_backbones_results_ScanObjectNN} report clean and robust accuracies for Vanilla, AT, and MAPR models, evaluated both with and without the SOR defense on ModelNet40 and ScanObjectNN, respectively. The final column reports the average robustness accuracy across all attacks, excluding clean accuracy.

\paragraph{Results on ModelNet40.}
On ModelNet40, MAPR substantially improves robustness over vanilla training across all five architectures. Without SOR, the average robustness accuracy improves from 37.90\% to 53.09\% for PointNet, from 42.59\% to 69.88\% for PointNet++, from 45.62\% to 65.67\% for DGCNN, from 66.21\% to 74.08\% for CurveNet, and from 42.90\% to 72.58\% for PointMLP. Averaged across architectures, this corresponds to a robustness gain of +20.02 percentage points over the vanilla baselines.

The largest improvements are observed for PointMLP (+29.68 p.p.), PointNet++ (+27.29 p.p.), and DGCNN (+20.05 p.p.), suggesting that MAPR is particularly effective for architectures whose learned representations are highly sensitive to local geometric perturbations. CurveNet shows a smaller but still positive gain (+7.87 p.p.), likely because its curve-based local aggregation already provides stronger geometric inductive bias than simpler architectures.

MAPR is also complementary to SOR. When combined with SOR, MAPR+SOR improves average robustness over Vanilla+SOR for all backbones, with particularly strong gains for PointNet++ and PointMLP. This suggests that MAPR is broadly compatible with input-level defenses.

\paragraph{Results on ScanObjectNN.}
We further evaluate MAPR on ScanObjectNN, a more realistic and challenging benchmark with scanned objects, clutter, occlusion, and acquisition artifacts. Compared with ModelNet40, absolute clean and robust accuracies are lower across all methods, reflecting the increased difficulty of the dataset.

Despite this harder setting, MAPR continues to improve robustness over vanilla training for all five architectures without SOR. Average robustness improves from 17.71\% to 22.80\% for PointNet, from 23.32\% to 37.48\% for PointNet++, from 26.18\% to 32.32\% for DGCNN, from 35.88\% to 42.64\% for CurveNet, and from 26.62\% to 38.60\% for PointMLP. Averaged across architectures, MAPR achieves a robustness gain of +8.83 percentage points over the vanilla baselines.

The gains on ScanObjectNN are smaller and less uniform than on ModelNet40, which is expected given the stronger real-world variability of the dataset. Nevertheless, MAPR improves clean accuracy for PointNet, DGCNN, CurveNet, and PointMLP. The main exception is PointNet++, where MAPR improves robustness substantially but reduces clean accuracy, indicating a stronger clean-robust trade-off for this backbone on real scanned data.

\paragraph{Comparison with adversarial training.}
Although MAPR does not use adversarial examples during training, it often matches or exceeds AT in robustness. On both datasets, MAPR achieves the best average robustness without SOR for PointNet, PointNet++, CurveNet, and PointMLP, while AT is slightly better for DGCNN. These results suggest that geometry-aware regularization can provide a competitive alternative to adversarial training, while remaining conceptually distinct and compatible with defense mechanisms.

\paragraph{Computational overhead.}
MAPR introduces intrinsic feature extraction and a second forward pass during training, increasing computational cost compared to vanilla training. Across architectures, MAPR is approximately 2-2.5$\times$ slower than vanilla training over 100 epochs. However, it remains substantially more efficient than PGD-based adversarial training, achieving roughly a 2$\times$ speedup on average while providing improved robustness across multiple attack settings.

\paragraph{Attack-specific behavior.}
MAPR improves robustness across a broad set of perturbations, but its behavior is attack-dependent. The most consistent gains appear under PGD-$\ell_2$, PGD-$\ell_\infty$, BIM, and SI-PGD, supporting the view that manifold-aligned regularization reduces sensitivity to perturbations that preserve local geometric structure. However, Add-100 remains challenging for some backbones, likely because the added adversarial points introduce local geometric patterns that do not correspond to the underlying object surface. In these cases, combining MAPR with SOR can mitigate the effect by removing isolated or abnormal points before classification.

\subsection{Ablation Studies}
\label{sec:ablations}

\paragraph{Effect of MAPR components.}
We evaluate the contribution of each component of MAPR---intrinsic features and Lipschitz regularization---to the overall robustness. Each model is evaluated under three configurations:
(1) \textit{Intrinsic-only}: the model includes intrinsic features concatenated with the $(x,y,z)$ coordinates but omits Lipschitz regularization;
(2) \textit{Lip-only}: the model uses only the $(x,y,z)$ coordinates with Lipschitz regularization; and
(3) \textit{Full MAPR}: the complete model combining both intrinsic features and Lipschitz regularization.
Results on ModelNet40 and ScanObjectNN are summarized in Tables~\ref{tab:ablation_mapr_components_MN40} and~\ref{tab:ablation_mapr_components_SObjNN}, respectively.

On ModelNet40, the full MAPR formulation achieves the highest average robustness for all five architectures. Compared with Intrinsic-only, Full MAPR improves average robustness by +13.97, +26.54, +21.56, +7.80, and +29.88 p.p. for PointNet, PointNet++, DGCNN, CurveNet, and PointMLP, respectively. Compared with Lip-only, Full MAPR improves average robustness by +10.62, +19.12, +13.01, +2.08, and +21.23 p.p., respectively. These results show that intrinsic features and Lipschitz regularization are complementary: each component provides useful geometric bias, but their combination produces the strongest robustness.

On ScanObjectNN, the same trend largely holds, although the gains are smaller due to the increased difficulty of real scanned data. Full MAPR achieves the best average robustness for every backbone. Compared with Intrinsic-only, it improves average robustness by +5.26, +14.14, +6.14, +7.23, and +15.28 p.p. for PointNet, PointNet++, DGCNN, CurveNet, and PointMLP, respectively. Compared with Lip-only, Full MAPR improves average robustness by +4.61, +12.16, +0.50, +2.80, and +4.73 p.p., respectively.

These results also clarify the behavior observed in Section~\ref{sec:main_results} for the Add-$k$ attack. On ModelNet40, the Lip-only component is highly effective for PointMLP under Add-100 but less stable for PointNet, DGCNN, and CurveNet. Full MAPR mitigates this instability in several cases, but Add-100 remains a difficult perturbation because it introduces optimized points that can form artificial local structures. On ScanObjectNN, Add-100 behavior is similarly architecture-dependent, reinforcing that this attack tests a different failure mode from smooth intrinsic perturbations.

\section{Conclusion}
\label{sec:conclusion}

In this paper, we investigated the geometric origins of adversarial fragility in point cloud classification. Standard classifiers are vulnerable to small perturbations of the input point cloud that often lead to incorrect predictions. We hypothesize that this fragility is closely tied to a geometric mismatch between the intrinsic structure of the object’s surface and the latent geometry learned by the model. Building on this idea, we developed a geometric framework for interpreting robustness in 3D, introducing a notion of manifold misalignment together with an intrinsic consistency regularization term designed to address it.

Guided by this theory, we introduced Manifold-Aligned Point Recognition (MAPR), a regularization framework that combines (1) intrinsic geometric features and (2) a local intrinsic consistency loss. Across five representative point cloud architectures and a diverse set of adversarial attacks, MAPR consistently improves robustness over vanilla models on both the ModelNet40 and ScanObjectNN datasets. In particular, MAPR achieves average robustness gains of +20.02 and +8.83 p.p. on ModelNet40 and ScanObjectNN, respectively, without relying on adversarial training or additional data. We further showed that MAPR is complementary to existing defenses such as SOR: in many cases, the MAPR+SOR combination yields even stronger robustness while maintaining competitive clean accuracy.

Overall, our results suggest that robustness in point cloud models is closely connected to the geometric structure of the learned latent space. By encouraging alignment between intrinsic surface geometry and latent representations, MAPR provides a simple and effective alternative to purely attack-driven defenses. We hope this work motivates future research on geometry-aware robustness methods and representation learning for 3D deep networks.

\appendix

\section{Implementation Details: Intrinsic Features}
\label{sec:appx_2}

We describe the implementation of the intrinsic features described in Sec.\ref{sec:method}.

\begin{itemize}
    \item We construct a $k$-NN graph using $k=20$ nearest neighbors. 
    \item Each edge is assigned a Gaussian weight
    \begin{equation}
        w_{ij} = \exp\!\left(
            -\frac{\|x_i - x_j\|^2}{\sigma_i^2}
        \right),
    \end{equation}
    where $\sigma_i$ is the distance from $x_i$ to its farthest neighbor in 
    $\mathcal{N}(x_i)$.  
    The adjacency matrix is symmetrized and row-normalized to obtain a 
    random-walk diffusion operator.
    \item We use four diffusion steps,
    \[
        t \in \{1, 2, 4, 8\},
    \]
    and compute both diffused coordinates $A^t X$ and a diffused density signal
    $A^t \mathbf{1}$ at each scale.
    \item Curvature is estimated using the random-walk Laplacian
    $L_{\mathrm{rw}} = I - A$, applied to the coordinates, and we take the
    resulting $\ell_2$ magnitude.
    \item The intrinsic descriptor concatenates all diffusion features
    (4 steps $\times$ 4 channels each) together with the curvature channel,
    yielding 17 intrinsic features per point.
    \item Including the original coordinates, the total number of input
    features is 20 per point (3 geometric coordinates + 17 intrinsic features).
\end{itemize}

\section{Effect of $\lambda_{\text{lip}}$}
\label{sec:appx_1}

The parameter $\lambda_{\text{lip}}$ controls the strength of the manifold-alignment regularization term (See Eq.~\ref{eq:total_loss}). As mentioned in Sec.~\ref{sec:experimental_setup}, the regularization weight increases linearly from 0 to $\lambda_{\text{max}}$ (epochs 1 to 15) and remains stable after that.

\begin{figure}
  \centering
  \includegraphics[width=\linewidth]{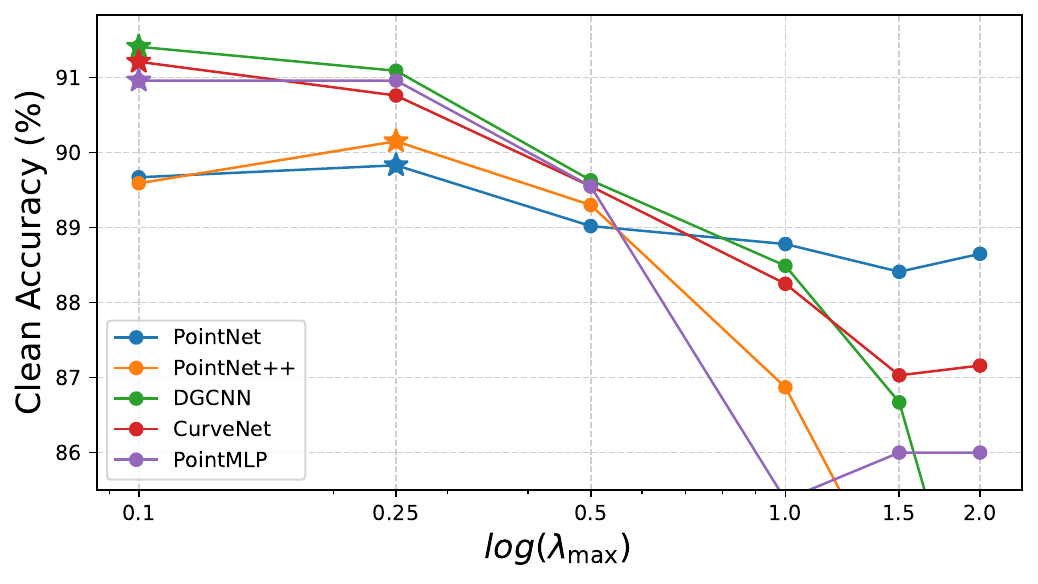}
  \caption{Clean accuracy of five 3D backbones under different MAPR regularization strengths $\lambda_{\text{lip}}$ for ModelNet40. Star markers indicate the best-performing setting for each backbone.}
  \label{fig:tests_lambda_max_MN40}
\end{figure}

\begin{figure}
  \centering
  \includegraphics[width=\linewidth]{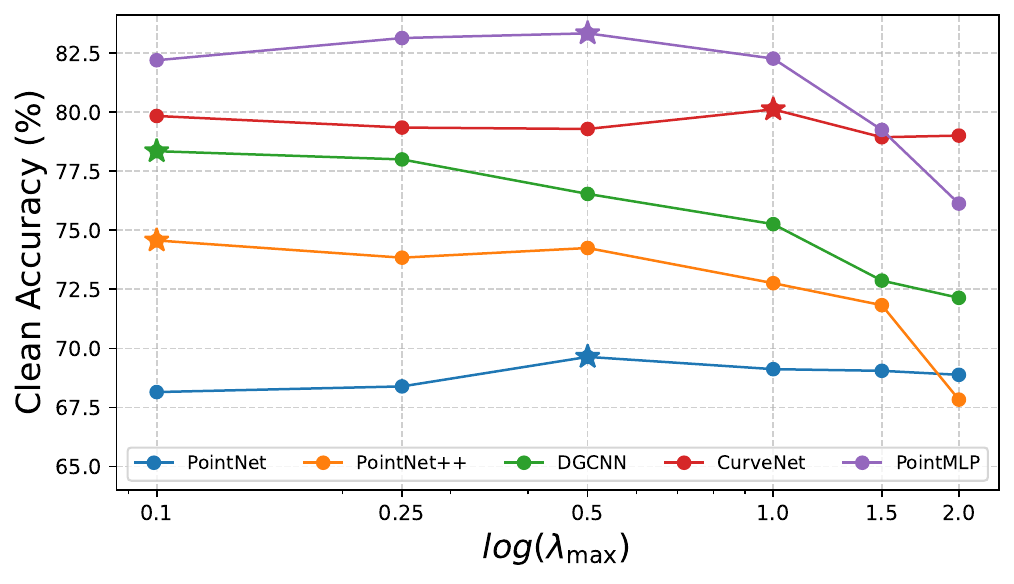}
  \caption{Clean accuracy of five 3D backbones under different MAPR regularization strengths $\lambda_{\text{lip}}$ for ScanObjectNN. Star markers indicate the best-performing setting for each backbone.}
  \label{fig:tests_lambda_max_SObjNN}
\end{figure}

Different architectures require different regularization strengths. To find the most appropriate value of $\lambda_{\text{max}}$ for each architecture, we evaluate performance across multiple $\lambda_{\text{max}}$ values ($0.1$, $0.25$, $0.5$, $1.0$, $1.5$, $2.0$) and choose the value that yields the highest classification accuracy. Results for ModelNet40 and ScanObjectNN are shown in Figs.~\ref{fig:tests_lambda_max_MN40} and ~\ref{fig:tests_lambda_max_SObjNN}, respectively.

\printcredits

\bibliographystyle{cas-model2-names}

\bibliography{main}





\end{document}